# A Nested Weighted Tchebycheff Multi-Objective Bayesian Optimization Approach for Flexibility of Unknown Utopia Estimation in Expensive Black-box Design Problems

Arpan Biswas[1,3], Claudio Fuentes[2] and Christopher Hoyle[1]*

[1]Department of Mechanical Engineering, Oregon State University, Corvallis, OR 97331 USA

[2]Department of Statistics, Oregon State University, Corvallis, OR 97331 USA

[3]Center for Nanophase Materials Sciences, Oak Ridge National Laboratory, Oak Ridge, TN 37831

We propose a *nested weighted Tchebycheff Multi*-objective Bayesian optimization (MOBO) framework where we build a regression model selection procedure from an ensemble of models, towards better estimation of the uncertain parameters (utopia) of the weighted-Tchebycheff expensive black-box multi-objective function. In existing work, a weighted Tchebycheff MOBO approach has been demonstrated which attempts to estimate the model parameters (utopia) in formulating acquisition function of the weighted Tchebycheff multi-objective black-box functions, through calibration using a priori selected regression model. However, the existing MOBO model lacks flexibility in selecting the appropriate regression models given the guided sampled data and therefore, can under-fit or over-fit as the iterations of the MOBO progress. This ultimately can reduce the overall MOBO performance. As it is too complex to a priori guarantee a best model in general, this motivates us to consider a portfolio of different families (simple-to-complex) of predictive models that have been fitted with current training data, guided by the WTB MOBO; the best model is selected following a user-defined prediction root mean-square-error-based approach. The proposed approach is implemented in optimizing a multi-modal benchmark problem and a thin tube design under constant loading of temperature and pressure, with minimizing the risk of creep-fatigue failure and design cost. Finally, the *nested weighted Tchebycheff MOBO* model performance is compared with different MOBO frameworks with respect to accuracy in parameter estimation, Pareto-optimal solutions and function evaluation cost. This method is generalized enough to consider different families of predictive models in the portfolio for best model selection, where the overall design architecture allows for solving any high-dimensional (multiple functions) complex black-box problems and can be extended to any other global criterion multi-objective optimization methods where prior knowledge of utopia is required.

*chris.hoyle@oregonstate.edu

*Keywords*: Multi-Objective Bayesian Optimization; Nested Design; Weighted Tchebycheff, Surrogate Modelling, Model Parameter Estimation.

1. **Introduction**

In the early design phase, it is very important for the designers to be able to identify potential good design decisions in a large design space while the design cost is low. In practice, most of the design problems are too complex to be handled by simple optimization frameworks due to having constraints in cost, time, formulation, etc. In many design optimization problems, it is difficult to numerically formulate an objective function and therefore we consider those problems as having black box objective functions with high function evaluation cost. When we have no or limited knowledge on the expensive true objective function, we cannot guarantee the maximization of our learning towards an optimal solution without proper guidance or expertise. Furthermore, due to the mentioned high function evaluation cost, exhaustive search is not a feasible option. In such black box engineering design problems, a Bayesian Optimization technique (BO), which eliminates the need of standard formulation of objective functions [1]–[3], is widely applied in sequential learning to provide better guided design sampling to minimize expensive function evaluations in finding the optimal region of the unknown design space.

*1.1. Research motivation*

In [4], [5], a design architecture to solve multi-objective black-box problems-- *weighted Tchebycheff multi-objective Bayesian optimization* (MOBO) is demonstrated, where the unknown utopia values (model parameters) are estimated iteratively using a priori selected predictive model. The utopia point is the optimum of each objective individually and is needed for the formulation of the global criterion multi-objective methods, such as the weighted Tchebycheff (WTB). The stated framework reduces the model complexities, by reducing the high-dimensional function space to a 1D multi-objective function space in formulating acquisition function, and also increases the overall Pareto-optimal solution accuracy from estimating utopia instead of considering an educative guess [6], [7]. However, this existing architecture lacks flexibility as a simple linear regression model has been pre-defined for utopia estimation at each iteration of the MOBO. Obviously, the simple regression model can under-fit if the unknown objective function is very complex (non-linear). To avoid this, on the other hand, predefining a complex predictive model can lead to overfitting if the unknown objective function is truly simpler. In both the cases, the error in the utopia estimation increases.

*chris.hoyle@oregonstate.edu

In the WTB, at a given vector of weighting factors $w$ on two objectives $f_1$ and $f_2$, the method forms a rectangle and searches along the diagonal to find the Pareto-optimal solution on the Pareto frontier curve. Let us assume, in Fig. 1, that the true Pareto-optimal solution lies at ($C$) where true utopia ($u$) is shown. If the location of ($u$) is unknown and is estimated to be at position ($u'$) (in X or Y), we see with the same weighting factor ($w$) that the WTB will now project the Pareto-optimal solution to $C'$ (the diagonal is shifted). Thus, we cannot find the true optimal solution ($C$) for weighting factor ($w$), if we use the incorrect estimate for the utopia of $u'$. In the domain of expensive black-box problems, we do not find the true utopia and need to settle with a cheap surrogate model to estimate the same; however, the goal would be to maximize the accuracy of estimation. With increasing the error in estimating utopia value, for instance due to inappropriate selection of predictive models, we can observe an increase in deviation of the optimal solutions from the desired trade-off between the objectives, thereby, the MOBO model may perform poorly. For obvious reasons, it is hard to know a priori what is the true nature of the black-box objective functions, thus challenging to a priori selecting an appropriate predictive model. Also in MOBO, as the data sampling is done sequentially towards finding the multi-objective optimal region, the performance of different simple-to-complex regression models can vary iteratively depending on the available data at each iteration, thereby exacerbating the challenges in appropriate handling of uncertain parameters in the weighted Tchebycheff MOBO architecture.

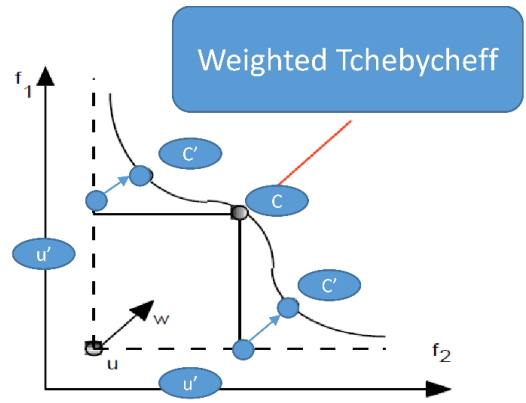

**Figure 1.** Incorrect utopia values leading to deviate from true pareto-optimal solution

*1.2. Research contribution*

To increase model performance, we introduce a predictive model selection approach nested into the existing weighted Tchebycheff MOBO. Then, with iteratively augmented training data in the MOBO, we estimate the utopia values from the iteratively selected cheap surrogate predictive models through this nested design. We call this the *nested weighted Tchebycheff*

*chris.hoyle@oregonstate.edu

*MOBO*. For simplicity in this paper, we also refer this as *nested MOBO*. Our main goal is to propose a design architecture for multi-objective black box problems in which one properly utilizes the existing sampled training data during model calibration and improves in predicting model uncertain parameters (e.g. utopia) rather than having a fixed pre-defined predictive model, by adding flexibility to choose the current best model from the portfolio of ensemble of pre-defined models. Another goal is to generalize the model comparison and selection approach where different families of predictive models can be put into consideration for building the portfolio. To illustrate the approach, we consider a numerical test problem and an engineering problem: *cyclic pressure-temperature loaded thin tube design* where we will be able to compare the results obtained from the proposed model with the true solutions from exhaustive search. However, it is to be noted the scope of the proposed *nested weighted Tchebycheff MOBO* is applicable to any multi-objective black-box problem, like in a diffusion bonded Compact Heat Exchanger [8].

The road-map of this paper is as follows: Section 2 provides an overview on single and multi-objective Bayesian optimization, multi-objective optimization methods, model calibration techniques. Section 3 provides the general description of the proposed *nested weighted Tchebycheff MOBO* or *nested MOBO* design architecture. Section 4 describes the cases studies with the implementation of the proposed design architecture. Section 5 illustrates the results of the *nested weighted Tchebycheff MOBO* among other design architectures for the case studies in terms of different performance parameters. Section 6 concludes the paper with final thoughts.

## 2. Overview on Bayesian Optimization

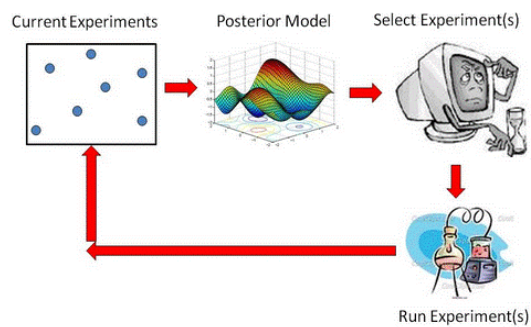

**Figure 2.** Bayesian Optimization Framework

Bayesian optimization [9] is an emerging field of study in sequential design methods. It is considered a low-cost global optimization tool for design problems having expensive black-box objective functions. The general idea of the Bayesian optimization is to emulate an expensive

4*chris.hoyle@oregonstate.edu

unknown design space and find the local and global optimal locations while reducing the cost of function evaluation from expensive high-fidelity models. This approach has been widely used in many machine learning problems [10]–[14]. However, attempts have been made when the response is discrete like in consumer choice modelling problems where the responses are in terms of user preference ( Chu and Ghahramani 2005; Brochu, Cora, and de Freitas 2010). The idea is to approximate the user preference discrete response function into continuous latent functions using Binomial-Probit model for two choices [16], [17] and polychotomous regression model for more than two choices where the user can state no preference [18].

### 2.1. Gaussian process model (GPM)

BO adopts a Bayesian perspective and assumes that there is a prior on the function; typically, we use a Gaussian process (GP) prior. The overall Bayesian optimization approach has two major components: A predictor or Gaussian process model, and the acquisition function. As shown in Figure 2, starting with a Gaussian prior, we build a posterior Gaussian process model, given the current available data from the expensive function evaluations. The surrogate GPM then predicts the outputs of any unsampled designs within the region of feasible design space. The uncertainty of these outputs near the observational data is small and increases as the unsampled designs are farther away from the observational data, thereby related to kriging models where the errors are not independent. In order to incorporate and quantify uncertainty of the experimental or training data, one approach is to use a nugget term in the predictor GPM, which is largely discussed in the literature. It has been found that the nugget provides a better solution and a computationally stable framework [19], [20] as it mitigates the ill-conditioning of correlation matrix. The technique is similar to Tikhonov regularisation where a small positive number $\vartheta$ is added on the main diagonal of the correlation matrix, which is called the nugget. Furthermore, GPM has also been implemented in high dimensional design space exploration [21] and big data problems [22], as an attempt to increase computational efficiency. As an alternative to GPs, random forest regression has been proposed as an expressive and flexible surrogate model in the context of sequential model-based algorithm configuration [23]. Although random forests are good interpolators in the sense that they output good predictions in the neighbourhood of training data, they are very poor extrapolators where the training data are far away [24]. This can lead to selecting redundant exploration (more experiments) in the non-interesting region as suggested by the acquisition function in the early iterations of the optimization, due to having additional prediction error of the region far away from the training data. This motivates us to consider the GP model in Bayesian framework. However, in multi-objective settings, the GP model still fits each objective or response variables independently, thus assuming that objectives are uncorrelated. A survey of different GP packages available for different coding languages such as MATLAB, R, Python can be found in [25].

*chris.hoyle@oregonstate.edu

### *2.2. Acquisition function (AF)*

The second major component in Bayesian optimization is the acquisition function whose goal is to strategically select the best design locations for future experimentation, defined from the posterior simulations which are obtained from the GP model. The acquisition function predicts the improvement in learning of sampling new designs, thereby, guiding the search for the optimum and bringing the sequential design into the BO. The improvement value depends on exploration (unexplored design spaces) and exploitation (region near high responses). Thus, the acquisition function gives a high value of improvement to the samples whose mean prediction is high, variance is high, or both. Thus, by maximizing the acquisition function, we strategically select design points which have the potential to have the optimal (maximum value of the unknown function) and gradually reduce the error to align with the true unknown function with iterations. Throughout the years, various formulations have been applied to define the acquisition functions. One such method is the *Probability of Improvement* (PI) [26] which is improvement based acquisition function. Jones in [27] notes that the performance of PI(·) "is truly impressive;… however, the difficulty is that the PI(·) method is extremely sensitive to the choice of the target. If the desired improvement is too small, the search will be highly local and will only move on to search globally after searching nearly exhaustively around the current best point. On the other hand, if the small-valued tolerance parameter $\varepsilon$ in PI(.) equation, as in [27], is set too high, the search will be excessively global, and the algorithm will be slow to fine-tune any promising solutions." Thus, the *Expected Improvement* (EI) acquisition function [9] is widely used over PI which is a trade-off between exploration and exploitation. Another acquisition function is the *Confidence bound criteria* (CB) introduced by Cox and John [28], where the selection of points is based on the upper or lower confidence bound of the predicted design surface for maximization or minimization problem respectively. In multi-objective BO problems, the *Expected Improvement Hyper-volume* (EIHV) acquisition functions have been modelled to provide better performance [29], [30]. However, to increase the computational efficiency, K. Yang et al. (2019) modified the EIHV acquisition function into the *Expected Hyper volume gradient-based* (EIHVG) acquisition function and proposed an efficient algorithm to calculate it. To reduce the computational cost, other acquisition functions like the *Max-value Entropy Search* (MESMO) [32] and *Predictive Entropy Search* (PESMO) [33] have been formulated. Abdolshah et al. (2019) proposed a multi-objective BO framework with preference over objectives on the basis of stability and attempting to focus on the Pareto front where preferred objectives are more stable. However, due to the computationally complexity of all these MO-BO approaches with increasing number of objectives, the weighted Tchebycheff method has been augmented in MO-BO in [35] with a ridge regularization term to smoothen the converted single multi-objective function. Another a priori method is ParEGO [36], where the weights are randomly assigned with a uniform distribution. Likewise, we [5] have implemented a weighted Tchebycheff method in MOBO with a priori user-defined weights and by introducing regression-based model calibration to estimate the parameter (utopia) of the multi-objective function. As stated earlier, the scope of this paper is to modify the architecture [5] in order to enhance the model performance.

*chris.hoyle@oregonstate.edu

### 2.3. Multi-Objective Optimization (MOO)

The numerical optimization problems can be classified on the basis of number of objective functions as Single (SOO) and Multi-Objective Optimization problems (MOO). MOO is the extension of SOO with having more than one objective. It is obvious that SOO is relatively simpler with lower computational cost; however, in practical problems it may be a challenge to formulate a single objective problem and therefore, much focus has been given on Multi-Objective Optimization methods (MOO). The question we want to solve in any MOO is the optimal design decisions under user defined preferences of objectives; optimal solutions at different trade-offs of objectives are represented by a *Pareto frontier*. The methods to solve MOO problems can be classified into *a priori* and *a posteriori* Methods. The most fundamental *a priori* method is the *Weighted Sum Method* (WS) [36] where we transform all the objectives into a single objective of weighted objectives. The method though simple and fast, is inefficient to find the true pareto-optimal points, mainly in the non-convex region. To increase the performance, a global criterion method, the *Weighted Tchebycheff method* (WTB) [37]–[39] is introduced where the multi-objectives are combined into a weighted distance metric that guarantees finding all the Pareto-optimal solution. Another a priori method is the $\epsilon$-*Constraint method* where the most critical objective is picked and treated as a single objective problem where the other objectives treat as constraints [40]. Another a priori method is the *Lexicographic method* where the objective functions are organized sequentially in order of preferences [41]. Some *a posteriori* methods used in MOO problems are *Vector Evaluated Genetic Algorithm* (VEGA) [42], *Niche Methods* [43], *Particle Swarm Algorithm* (PSO) [44], *NGSA 2* [45], etc. Readers can view [36], [46] for the application in practical multi-objectives design problems using the above mentioned priori and posteriori methods, as well as ref. [47] for additional methods. It is a challenging task to pick a best method, as the performance of the methods depend on the problems and its constraints on dimension, formulation, computational cost, uncertainty in design etc. In this paper, we choose to focus on *Weighted Tchebycheff* method in the MOBO framework, and extend it to a nested architecture.

*chris.hoyle@oregonstate.edu

### 2.4. Model Calibration and Validation

In the data-driven modelling approach, the performance of the model can be improved by tuning the unknown model parameters based on how closely they represent the real complex expensive designs. This tuning of model parameters is *model calibration* while the measurement of the performance of the model after calibration, by comparing the model solution with the true solution, is the *model validation*. Adequate true data is desired for model calibration and validation; however, getting true data can be very expensive and sometimes not possible in complex design problems. Due to the use of true data in model calibration and validation, it is necessary for the designer to know which parameters are likely to be more sensitive to the model before they start collecting true data. Calibrating and validating a model for any insensitive (minimal sensitive) model parameters will multiply the cost of true data collection. Thus, a sensitivity analysis should be done between the model parameters to observe the relationship between input and output functions [49], [50] by conducting a study with limited existing true data, or approximated model data, or solely on the knowledge of the designers. Regression analysis is one of the low-cost statistical tools for sensitivity analysis which gives the variation of the output to the inputs. Finally, assuming true data from expensive evaluations, model calibration and validation can be done iteratively until designers find the best model parameter values, satisfying the model overall performance in terms of the trade-off between model accuracy and cost of data sampling. Model calibration and validation becomes further challenging when we have uncertainty on the model parameters and/or the true output values [51]. In such cases, a distribution of the model parameter is considered instead, and a distribution of the model output is validated with true values (if the true outputs are known and fixed) or the true distribution of the outputs (if the true outputs are known but uncertain). When we do not know the true distribution of the uncertain outputs, model validation is challenging and expert knowledge is required. Reviews of different techniques for model calibration and validation, considering the availability of true data and the model input-output uncertainty is provided in these papers [52].

*chris.hoyle@oregonstate.edu

## 3. Design Methodology

In this section, we present the design methodology of the proposed nested weighted Tchebycheff Multi-objective Bayesian optimization (MOBO).

### 3.1. Outer Loop: Weighted Tchebycheff MOBO

The outer loop of the proposed design architecture defines the Bayesian optimization, solving for multiple objectives, using the weighted Tchebycheff based acquisition function.

The weighted Tchebycheff multi-objective black-box function in our Bayesian optimization setting for either maximization or minimization of the multiple objectives is as follows:

$$\min_{X} Y_{multi} = \min_{X} \max_{i \in N} \{w_i | \hat{\mu}(Y_i | \mathbf{\Delta_k}) - \widehat{\mu_k}(u_i) | \} \tag{1}$$

where $w_i$ is the weighting factor of $i^{th}$ objective; $N = 2$ is the total number of objectives; $\hat{\mu}(Y_i | \mathbf{\Delta_k})$ is the estimated mean value of the $i^{th}$ objective function, given the posterior surrogate Gaussian process model $\mathbf{\Delta_k}$, at iteration $k$ of the MO-BO; $\widehat{\mu_k}(u_i)$ is the estimated mean utopia value $(u_i)$ of the $i^{th}$ objective function where , which has been calibrated from the selected regression model. Minimizing the maximum weighted distance from the utopia among the objective functions will provide the non-dominated solutions or pareto-optimal solutions. It can be easily transformed into maximizing the acquisition function of $Y_{multi}$, as $\max_{X} U(-Y_{multi})$, thereby selecting samples for expensive function evaluations with higher likelihood of being a Pareto-optimal solution. Here, we reduce the high-dimensional function space into a single multi-objective function (1D space) using weighted Tchebycheff method and then maximize with any single objective based acquisition function. Thus, we minimize the complexity arises from curse of dimensionality. As stated earlier, the outer loop architecture is similar to our existing work, however, for the convenience we provide the high level algorithm 1 of the outer loop.

**Algorithm 1:** *Outer loop of nested weighted Tchebycheff MOBO*:

1. Define the design space or the region of interest for the given problem.
    1.1. From the defined design space, generate a grid matrix using a DOE approach.
    1.2. Build a training data matrix with the starting sampled data. Create the training data matrix, assuming at iteration k, $\boldsymbol{D_k} = [\boldsymbol{X_k}, \boldsymbol{Y(X_k)}]$.
2. Validate the design with constraints (if any). Filtered with only feasible starting sampled data, $\boldsymbol{D_{f,k}} = [\boldsymbol{X_k}, \boldsymbol{Y(X_k)}]$ and $\boldsymbol{D_{f,k}} \in \boldsymbol{D_k}$. Let $\boldsymbol{X_f}$ be the feasible un-sampled grid designs.

*chris.hoyle@oregonstate.edu

3. Predicting the utopia from Regression Analysis, given $D_{f,k}$. *This step will run the inner-loop regression model selection.*

4. Develop posterior surrogate Gaussian Process models, $\Delta_k$, for each objective independently, given $D_k$.

5. Calculate estimated means and MSE of the designs, $X_f$, forming two matrices $\hat{\mu}(Y_f|\Delta_k)$ and $\hat{\sigma}^2(Y_f|\Delta_k)$ respectively.

6. Define the weighted Tchebycheff multi-objective Acquisition Function and maximize the acquisition function, $\max U(-Y_{multi}|\Delta_k)$. Augment the data with best design for evaluation and the respective estimated mean, $\mathcal{D}_k = [D_k; [X_{f,max}, \hat{\mu}(Y(X_{f,max})|\Delta_k)]$ where $X_{f,max} \in X_f$.

7. Check for convergence criteria 1. If not met, run $j = 1:n$ of Steps 4-7. Select the best n design locations, $X_{f,max} = [X_{f,max,1}; \ldots; X_{f,max,n}]$, to proceed to the next round of expensive function evaluations.

   - Convergence Criteria 1:

   $$if\ j = 1, \max U(-Y_{multi}|\Delta_k) \leq \alpha$$

8. Conduct expensive function evaluations. New experimental data is $[X_{k+1}, Y(X_{k+1})]$ where $X_{k+1} = X_{f,max}$.

9. Augment training data matrices for regression model (Step 3) and GP model (Step 4) as $D_{f,k+1} = [D_{f,k}; [X_{k+1}, Y(X_{k+1})]$ and $D_{k+1} = [D_k; [X_{k+1}, Y(X_{k+1})]]$. Repeat Steps 3-9 until convergence.

10. If convergence criteria 2 met, stop the MOBO model and note the optimal solutions.

    - Convergence Criteria 2: Reaching maximum number of expensive function evaluations.

### *3.2. Inner Loop: Model Selection for unknown parameter estimation*

Here, we describe broadly the model selection algorithm to estimate utopia values following a prediction root mean square error approach, which is nested in the weighted Tchebycheff MOBO (Step 3, Algorithm 1). We will call this the inner loop of the weighted Tchebycheff MOBO. Figure 3 shows the detailed flow chart of the model selection procedure. The key point in the procedure is to compute two criteria, which together define the model selection criterion. In Fig. 3, the steps (blocks) highlighted in blue, green and red are involved in criteria 1, criteria 2 and both criteria 1 and 2 respectively. The selection procedure is based upon the research objective to answer how good a certain model has estimated the certain parameters (utopia values); we do not consider the complexity of the model as part of the selection criteria. Our work aims to add flexibility to the model comparison among different families of models. In addition to classical linear regression models we also consider SVMR and GPM, together with Bayesian linear regression models for estimation methods. The comparison ignores the trade-off between model accuracy and computational cost due to the increase of model parameters (e.g., number of regression co-efficient): this is because in the proposed BO framework, the computational cost for regression model fitting is negligible compared to the function evaluations from any expensive black-box design problems. It is to be noted though our case studies described are not "actual" but rather assumed expensive for the sake of the proof of concept; the method presented in this paper is intended for solving any expensive black-box designs.

*chris.hoyle@oregonstate.edu

### 3.2.1. *Criteria 1: Global improvement*

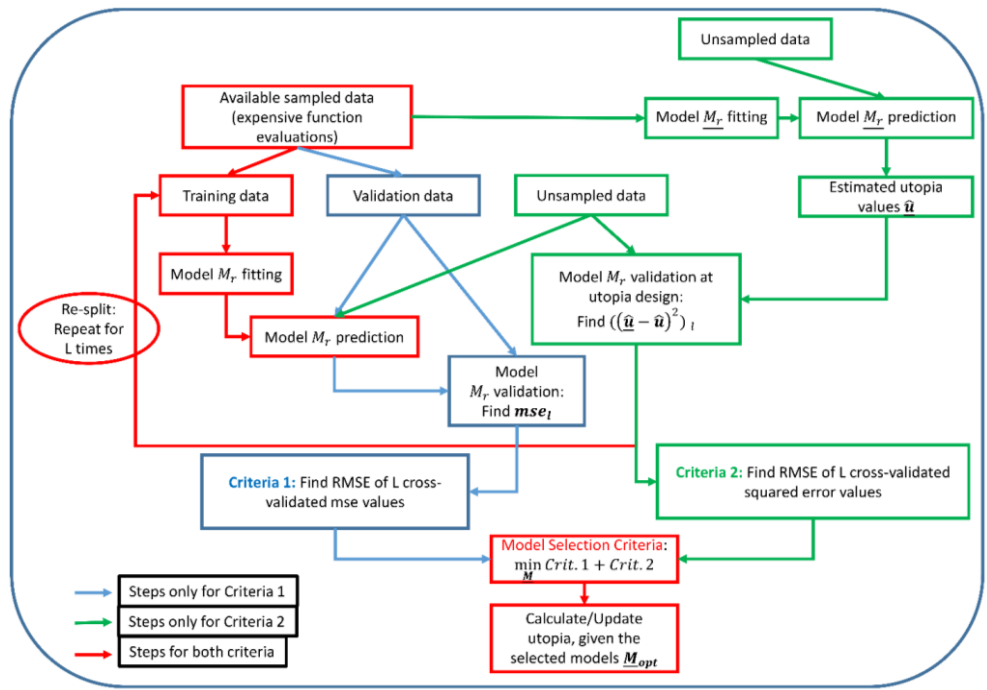

**Figure 3**. Model selection algorithm to estimate utopia (Inner loop of the MO-BO)

The first criterion to consider in the model selection procedure is the overall global improvement by the model to estimate the designs across the feasible design space. Thus,

*chris.hoyle@oregonstate.edu

criteria 1 is focused on selecting a model at iteration $k$ of the nested MOBO model, which gives the best fit in general. With a best fit in general, the model has higher likelihood to provide better estimation of utopia values. Algorithm 2a describes the steps to follow for computing criteria 1:

**Algorithm 2a:** *Inner loop of nested weighted Tchebycheff MOBO: Computing Criteria 1:*

*Step 1: Training Data:* We define $X$ as design input variables and $Y$ as output functions. Create the feasible sampled data matrix, assuming at iteration $k$ of the MO-BO, $\boldsymbol{D}_{f,k} = [\boldsymbol{X}_k, Y(\boldsymbol{X}_k)]$ and $\boldsymbol{D}_{f,k} \in \boldsymbol{D}_k$. Define the ensemble of models to estimate utopia as $M$, which build the model portfolio. The portfolio of models for each utopia can be different, thus adding flexibility to the choice of models for different functions.

*Step 2: Conduct Monte-Carlo cross-validation:*

   Step 2a. Split $\boldsymbol{D}_{f,k}$ into two subsets of training $\boldsymbol{D}_{f,k}^T$ and validation $\boldsymbol{D}_{f,k}^V$ datasets without replacement at the proportion of $p^T$ and $p^V$ respectively where $p^T + p^V = 1$.

   Step 2b. Given $\boldsymbol{D}_{f,k}^T$, fit all the pre-defined regression models to estimate objective $i$, $\mathbf{M}_{r,i,k}$, where $r$ is the defined regression model number.

   Step 2c. Estimate the objectives and create vectors of $\widehat{\mu_k}(Y_i)$ for each input design in $\boldsymbol{D}_{f,k}^V$, given the regression model $\mathbf{M}_{r,i,k}$.

   Step 2d. Validate the estimated objectives in 2c with the true objective values in $\boldsymbol{D}_{f,k}^V$. Thus, calculate mean-squared-error $\epsilon_{1,i,l}$ to estimate objective $i$ and at iteration $l$ of the cross-validation for all the input designs $X_k^V$ in $\boldsymbol{D}_{f,k}^V$.

$$\epsilon_{1,i,l} = \mu(Y_i(X_k^V) - \widehat{\mu_k}(Y_i(X_k^V)))^2 \quad (2)$$

   Step 2e. Repeat Step 2a. - 2d. for $L$ times. In this case study, $L = 100$.

*Step 3: Define Criteria 1:* Calculate root-mean-square of the vectors MC cross-validated mean-squared-errors of $i^{th}$ objective, $\boldsymbol{\epsilon_{1,i,.}}$. Thus criteria 1 at iteration $k$ of MO-BO for $i^{th}$ objective can be stated as:

$$\mathrm{E}_{1,i,k} = \sqrt{\mu(\boldsymbol{\epsilon_{1,i,.}})} \quad (3)$$

### 3.2.2. Criteria 2: Local improvement

The second criterion is the local improvement by the model specific to our region of interest in the design space, which is the utopia region. It is to be noted that the purpose of the regression model is to have a good estimate at the utopia region, as a large error in any other region is not going to impact the MO-BO model performance. Although selecting a model with a good overall prediction accuracy as in criteria 1 has a likelihood of better estimation of utopia region as well, it does not provide a guarantee of the best estimation among other models. In other words, comparing fitted model 1 and 2 as in figs. 4a using the training data (blue dots), although model 1 has higher error for the validation data (green dots), it has lower error in predicting utopia design (red dot). In this example, we see the overall better fit model 2 has high error at the utopia region, which is the region of interest. Thus, along with the global improvement, we have also focused on the second criteria on reducing the estimation error specifically at the utopia region. However, doing so, the challenges lies that since the utopia design is unknown (red dot in figs. 4a, 4b), we will not know the true values of the objectives. This restricts the straightforward MC cross-validation as in section 3.2.1.

*chris.hoyle@oregonstate.edu

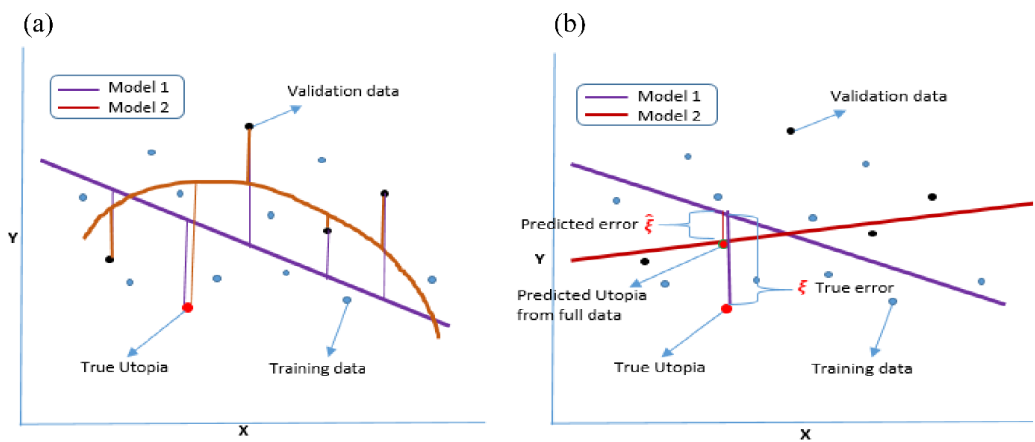

**Figure 4**. (a) Model Comparison: Model 1 has higher error on predicting validation data, but lower error on predicting utopia. Here model 1 and 2 are different regression models. (b) Objective for Criteria 2: To minimize the error $\hat{\xi}$ of predicted utopia between model 2 (fitted with all available data) and model 1 (fitted with only subsampled training data). Here, model 1 and 2 are same regression model, but since model 1 is fitted with more data (more knowledge), it is assumed to be more likely to get closer towards true utopia.

To mitigate this issue, we follow the assumption that fitting a model with more data will have higher likelihood for better estimation. Thus, the estimation error of the model, after fitting with the full feasible sampled dataset $D_{f,k}$ is likely to be lower than that when fitted with feasible subsampled dataset $D_{f,k}^T$. Following this, we assume that the reference utopia values for cross-validation are the estimated utopia values fit with the full feasible sampled dataset $D_{f,k}$ (denoted by red dot on model 2 regression line in fig. 4b) and the criteria 2 is to select the model which minimize the predicted error $\hat{\xi}$ in fig. 4b. Algorithm 2b describes the steps to follow for computing criteria 1:

*chris.hoyle@oregonstate.edu

**Algorithm 2b:** *Inner loop of nested weighted Tchebycheff MOBO: Computing Criteria 2:*

*Step 1: Training Data:* Same as stated in step 1, algorithm 2a. In addition, let $X_f$ be the feasible unsampled grid matrix for which the objective values are unknown.

*Step 2: Estimate utopia with full data:*

    Step 2a. Given $D_{f,k}$, fit all the pre-defined regression models to estimate objective $i$, $\mathbf{M}_{r,i,k}$, where $r$ is the defined regression model number.

    Step 2b. Estimate the objectives and create vectors of $\widehat{\underline{\mu_k}}(Y_i)$ for each unsampled design $X_f$, given the regression model $\mathbf{M}_{r,i,k}$.

    Step 2c. Estimate the utopia of objective $i$, as the minimum of vectors $\widehat{\underline{\mu_k}}(Y_i)$:

$$\widehat{\underline{\mu_k}}(u_i) = \min_{R_f, L_f, t_f} (\widehat{\underline{\mu_k}}(Y_i)) \tag{4}$$

Store the estimated utopia $\widehat{\underline{\mu_k}}(u_i)$ and the respective design values, $\underline{x_{f,i,k}} \in X_f$.

*Step 3: Conduct Monte-Carlo cross-validation:*

    Step 3a. Consider the same training $D_{f,k}^T$ subsampled dataset and the fitted model $\mathbf{M}_{r,i,k}$ as in step 2a and 2b, algorithm 2a.

    Step 3b. Estimate the objective or utopia values $\widehat{\underline{\mu_k}}(u_i)$ for respective input design $\underline{x_{f,i,k}}$, given the regression model $\mathbf{M}_{r,i,k}$.

    Step 3c. Validate the estimated utopia values in 3b with the estimation done with full samples in 2c. Thus, calculate squared-error $\epsilon_{2,i,l}$ to estimate utopia of objective $i$ and at iteration $l$ of the cross-validation.

$$\epsilon_{2,i,l} = (\widehat{\underline{\mu_k}}(u_i) - \widehat{\underline{\mu_k}}(u_i))^2 \tag{5}$$

    Step 3d. Repeat Step 3a. – 3c. for $L$ times. In this case study, $L = 100$.

*Step 4: Define Criteria 2:* Calculate root-mean-square of the MC cross-validated squared-errors of $i^{th}$ objective, $\epsilon_{2,i,\cdot}$. Thus the criteria 2 at iteration $k$ of MO-BO $i^{th}$ objective can be stated as,

$$E_{2,i,k} = \sqrt{\mu(\epsilon_{2,i,\cdot})} \tag{6}$$

Finally, the combined model selection criteria (among $r$ models) for $i^{th}$ objective to estimate utopia in the minimum of the addition of normalized values of eqns. 3 and 6 and can be stated as,

$$\min_{\underline{\mathbf{M}}_{i,k}} |E_{1,i,k}| + |E_{2,i,k}| \tag{7}$$

where the optimal estimated solution is $\widehat{\underline{\mu_k}}(u_i)_{opt} = \widehat{\underline{\mu_k}}(u_i|\underline{\mathbf{M}}_{i,k,opt})$. This value is inputed into the weighted Tchebycheff black-box objective function (eqn. 9) for MO-BO model calibration.

*chris.hoyle@oregonstate.edu

## 4. Case Studies

In this section, we describe one benchmark problems and one engineering design problem:

- *2-D Six-hump camel back* function and *Inversed-Ackley's Path* function
- *Thin tube design problem*

We introduce our implementation and illustration of the results of the proposed nested weighted Tchebycheff MOBO to these problems, to aid in understanding the predictive model selection approach, towards better estimation of the unknown parameters in the proposed surrogate or meta-model based design architecture (MOBO) for solving expensive/complex black-box multi-objective problems.

### *4.1. Benchmark problem*

For the benchmark problem, we considered two popular global optimization test functions as our objectives: the *2-D Six-hump camel back* function and *Inversed-Ackley's Path* function [53]. These functions are drawn from the literature on evolutionary algorithms and global optimization, and were chosen in this paper due to their non-linearity and multi-modal nature and thereby having a reasonable complexity in the guided search towards the optimal solutions. We assume these as expensive functions in our MOBO architecture. Below is the formulation of the 2D benchmark problem:

*Maximize Objective1: 2-D Six-hump camel back* function:

$$\max_{x_1, x_2} Y_1 = \max_{x_1, x_2} \left(4 - 2.1 x_1^2 + x_1^{\frac{4}{3}}\right) x_1^2 + x_1 x_2 + (-4 + 4x_2^2) x_2^2 \tag{8}$$

*Maximize Objective2: 2-D Inversed-Ackley's Path* function:

$$\max_{x_1, x_2} Y_2 = \max_{x_1, x_2} a\exp\left(-b\sqrt{\frac{x_1^2 + x_2^2}{2}}\right) + \exp\left(\frac{cos(cx_1) + cos(cx_2)}{2}\right) - a - \exp(1) \tag{9}$$

where $a = 20; b = 0.2; c = 2\pi; -3 \leq x_1 < 3; -2 \leq x_2 < 2; x_1, x_2 \neq 0$.

The values of a, b and c are considered as in [53].

### *4.2. Cyclic pressure-temperature loaded thin tube design problem*

For the design problem, we use the design of a thin tube under pressure-temperature cycling. As the tube is assumed to undergo constant loading of temperature and pressure, there will be risk of creep-fatigue failure which will vary with the design geometry. Fatigue damage is defined as cycling a test specimen at some fixed stress amplitude for enough cycles that it will develop micro-structural damage and eventually fail. Creep damage is defined as holding a test specimen at a fixed load for a long enough time that eventually it will develop micro-structural damage and fail. Creep-fatigue damage is therefore to do both of these simultaneously (e.g., stress-controlled cycle with a hold) such that the specimen will generally fail sooner than conducting the cycling or holding individually. The stated risk is defined based upon the location of the structural design in strain-defined regions such as elastic, shakedown, plastic or ratcheting. A similar example has been provided for thin tubes where the location of the designs can be numerically represented from a Bree Diagram [54] (Fig. Appendix A.1) in terms of pressure and thermal stresses. Under cyclic loading, the elastic and shakedown region in the Bree diagram are considered as the safe region where no strain accumulation occurs or the growth of residual strain is practically diminishing when sufficient loading cycles are applied. However, plastic and ratcheting in the Bree diagram are unsafe designs where the plastic strain accumulates until failure. Along with the risk, we have also considered the material cost of the tube. For a better tube design, we should minimize both the risk and cost functions. When the complexity of the design increases like in a diffusion bonded *Compact Heat Exchanger* [8], we cannot provide a numerical representation of such functions which define the location of designs. Also, considering different costs in the entire design process like material, component, manufacturing, assembly and quality checking makes the cost function hard to formulate. This makes both the objectives such as risk and cost functions to be black-box functions, representing the problem as black-box multi-objective design problem for which we proposed the Bayesian framework in the paper. Here, we assume the risk and cost functions for our case study problem are expensive black-box functions. We choose the design variables as radius (R), length (L), and thickness (t) of the tube. Next, we state the experimental procedure for the formulation of the multi-objective functions and the constraints for the thin tube problem.

*chris.hoyle@oregonstate.edu

### 4.2.1. *Experimental procedure*

To formulate the objective of risk of creep-fatigue failure, we need to find the location of any design in terms of elastic, plastic, shakedown and ratcheting, and the respective strain accumulation. We represent these outputs as the responses from the expensive experiments. In this paper, we consider the Bree diagram for a non-work-hardened material whose yield stress remains unchanged by changes in mean temperature as provided in Appendix A.1. For the sake of simplicity, we have ignored the further division of shakedown (S1, S2) and ratcheting (R1, R2) as shown in the figure, and assumed a single region of shakedown (S) and ratcheting (R), because the design risks are equivalent in the S1 and S2, and R1 and R2 regions, respectively. The three major steps we follow in the procedure are 1) Calculate Pressure and Temperature Stress of the design point, 2) Determine the region of the design in terms of elastic, plastic, shakedown and ratcheting and 3) Calculate the strain accumulation based on the location of the design. The detailed computation of the whole process for the thin tube can be found in [55].

*chris.hoyle@oregonstate.edu

### 4.2.2. Problem formulation of thin tube design

The objective functions and the constraints of the thin tube problem is as follows:

$$\min_{R,L,t} Y_1 = \min_{R,L,t} f(\psi, \xi, k) \tag{10}$$

$$\min_{R,L,t} Y_2 = \min_{R,L,t} P\rho\pi t L(2R - t) \tag{11}$$

Subject to (Constraints):

$$P\big((\hat{\mu}(Y_1(R,L,t)|\Lambda)) \leq 0.5\big) \geq R \tag{12}$$

$$\frac{L_D}{2\pi R t} - \sigma_y \leq 0 \tag{13}$$

$$L_D - \frac{\pi^3 E R^3 t}{4L^2} \leq 0 \tag{14}$$

$$\frac{R}{L} \leq \delta \tag{15}$$

Equation (10) is a distance function which measures the risk of creep-fatigue failure where $\psi \in \Psi$ is the region (elastic, shakedown, plastic and ratchetting) within the design space $\Psi$; $\xi$ is the total strain accumulation; $k$ is the $k^{th}$ iteration of the BO model. It is to be noted that in each iteration, with more experimental or training data (increase prior knowledge), the distance value for all the training data is re-evaluated. Equation (11) is the cost function where $P$ is the material cost per kg; $\rho$ is the density of the material. Equation (12) is the probabilistic constraint of creep-fatigue failure with Reliability factor, $R = 0.99$ where $\hat{\mu}(Y_1(R,L,t)|\Lambda)$ is the estimated mean of output objective, distance value, for the input design variables, given the converged posterior surrogate Gaussian Process model $\Lambda$. This posterior surrogate Gaussian Process model $\Lambda$ is first build and converged with a Bayesian optimization workflow as in [55]. The value 0.5 is the threshold since we set this distance value at the transition boundary line between safe and unsafe region. For details, the detailed formulation of distance metrics, which builds equations. 10 and 12, are in [55]. Equations 13-15 are the deterministic constraint equations for normal stress, buckling load and aspect ratio respectively where $L_D = 1KN$ is the load exerted on the wall of the thin tube, $\sigma_y = 205 MPa$ is the yield stress, $E = 207 GPa$ is the Young's modulus, $\delta = 0.025$ in this case study.

### *4.3. Description of the Model Portfolio:*

In this section, we focus on the ensemble of simple-to-complex regression models (building model portfolio), considered in this case study. Since the true nature of the design space or unknown objectives is assumed unknown (black-box), we have considered an ensemble of different flavours of regression models to understand their selection as the iteration of the MOBO progress. In total, we have considered 7 models to build the portfolio of models, used in regression analysis for estimation and prediction of outputs for given independent variables, including: 1. Mean model (MM), 2. Multiple linear regression model (MLR), 3. Log-transform of multiple linear regression model (log-MLR), 4. Bayesian multiple linear regression model (BMLR), 5. Second order polynomial model (SOP) (or quadratic model), 6. Support Vector Machine regression model (SVMR) and 7. Gaussian Process model (GPM). As previously stated, the design architecture is not constrained to implement only these models, and any regression models can be opted out or introduced into the nested MOBO framework at the start of the optimization (with any prior educative guess from existing similar problems and when the knowledge is very limited) or at the mid or later stage of the optimization (when we have better knowledge). However, for the sake of simplicity in model comparison, we have considered these models throughout the optimization process, assuming we have limited knowledge on the nature of the objectives to start with. Next, we have presented the formulation of fitting these models, using the sampled data in our case study. As our goal is to predict the utopia which is the optimal solution of an objective function independent of other objectives, we have considered $n = 2$ regression models for learning $n = 2$ objectives. The formulation of the models has been stated for the higher dimensional (3D) thin tube design; however, for the 2D benchmark problem (eqn. 8-9), the design variables $[R, L, t]$ can be replaced with $[x_1, x_2]$ in the following equations as required.

*chris.hoyle@oregonstate.edu

*Model 1:* The mean model (MM) is the simplest, only having intercepts in the regression model in equation (16-17):

$$\widehat{\mu_k}(Y_1|R, L, t) = \hat{\beta}_{0,1,k} \tag{16}$$

$$\widehat{\mu_k}(Y_2|R, L, t) = \hat{\beta}_{0,2,k} \tag{17}$$

where $\hat{\beta}_{0,1,k}$ ; $\hat{\beta}_{0,2,k}$ are the estimated regression coefficient (intercepts) for objective Eqn. (10) and (11), respectively, at iteration $k$ of the MO-BO model; $\hat{\mu}(.|R, L, t)$ is the estimated mean of the objective function, given the sampled data from expensive function evaluations in the Bayesian optimization framework.

*Models 2 & 3*: Likewise, the equations for the estimated means of the objective functions fitted in MLR and log-MLR models are as follows:

$$\widehat{\mu_k}(Y_1|R, L, t) = \hat{\beta}_{0,1,k} + \hat{\beta}_{1,1,k}R + \hat{\beta}_{2,1,k}L + \hat{\beta}_{3,1,k}t \tag{18}$$

$$\widehat{\mu_k}(Y_2|R, L, t) = \hat{\beta}_{0,2,k} + \hat{\beta}_{1,2,k}R + \hat{\beta}_{2,2,k}L + \hat{\beta}_{3,2,k}t \tag{19}$$

$$\widehat{\mu_k}(Y_1|R, L, t) = e^{\hat{\beta}_{0,1,k}} \times e^{\hat{\beta}_{1,1,k} \log R} \times e^{\hat{\beta}_{2,1,k} \log L} \times e^{\hat{\beta}_{3,1,k} \log t} \tag{20}$$

$$\widehat{\mu_k}(Y_2|R, L, t) = e^{\hat{\beta}_{0,2,k}} \times e^{\hat{\beta}_{1,2,k} \log R} \times e^{\hat{\beta}_{2,2,k} \log L} \times e^{\hat{\beta}_{3,2,k} \log t} \tag{21}$$

*chris.hoyle@oregonstate.edu

where $\widehat{\boldsymbol{\beta}}_{.,1,k} = (\hat{\beta}_{0,1,k}, \hat{\beta}_{1,1,k}, \hat{\beta}_{2,1,k}, \hat{\beta}_{3,1,k})^T$; $\widehat{\boldsymbol{\beta}}_{.,2,k} = (\hat{\beta}_{0,2,k}, \hat{\beta}_{1,2,k}, \hat{\beta}_{2,2,k}, \hat{\beta}_{3,2,k})^T$ are the vectors of estimated regression co-efficient (intercepts, slopes) for objective Eqn. (10) and (11) respectively at iteration $k$ of the MO-BO model. We considered full additive models, where both the objectives are a function of all predictor design variables (R, L and t). It is also to be noted that for the benchmark problem, the log-MLR models are avoided due to having negative values for the objective 2 (eqn. 9).

*Model 4*: In the Bayesian approach [56], [57], the estimation of regression coefficients for $i^{th}$ objectives and $k^{th}$ iteration of the MO-BO model, $\widehat{\boldsymbol{\beta}}_{.,i,k} = (\hat{\beta}_{0,i,k}, \hat{\beta}_{1,i,k}, \hat{\beta}_{2,i,k}, \hat{\beta}_{3,i,k})^T$ are complex as we need to formulate the prior $p(\boldsymbol{\beta})$, likelihood function $\ell(Y|X, \boldsymbol{\beta})$, and posterior distribution $p(\boldsymbol{\beta}|X, Y)$. For BMLR, we set Gaussian (or normal) priors as eqs. 22, 23.

$$p(\boldsymbol{\beta}_{.,1,k}) \sim MVN(\widehat{\boldsymbol{\mu}}_{\beta_{.,1,k-1}}, \widehat{cov}^2_{\beta_{.,1,k-1}} \boldsymbol{I}_{p+1}) \tag{22}$$

$$p(\boldsymbol{\beta}_{.,2,k}) \sim MVN(\widehat{\boldsymbol{\mu}}_{\beta_{.,2,k-1}}, \widehat{cov}^2_{\beta_{.,2,k-1}} \boldsymbol{I}_{p+1}) \tag{23}$$

Therefore, the posterior distribution is defined as equations (24, 25):

$$p(\boldsymbol{\beta}_{.,1,k}|Y_1, R, L, t) \propto l_k(Y_1|R, L, t, \boldsymbol{\beta}_{.,1,k}) \; p(\boldsymbol{\beta}_{.,1,k}) \tag{24}$$

$$p(\boldsymbol{\beta}_{.,2,k}|Y_2, R, L, t) \propto l_k(Y_2|R, L, t, \boldsymbol{\beta}_{.,2,k}) \; p(\boldsymbol{\beta}_{.,2,k}) \tag{25}$$

where $\widehat{\boldsymbol{\mu}}_{\beta_{.,1,k-1}}, \widehat{\boldsymbol{\mu}}_{\beta_{.,2,k-1}}, \widehat{cov}^2_{\beta_{.,1,k-1}} \boldsymbol{I}_{p+1}, \widehat{cov}^2_{\beta_{.,2,k-1}} \boldsymbol{I}_{p+1}$ are the estimated means and the variances, respectively, of the beta parameters for objectives 1 and 2 at iteration $k-1$ of the MO-BO model. $\boldsymbol{I}_{p+1}$ is the $(p+1)$ identity matrix where $p$ is the number of input design variables. Thus, the general idea is to compute the posterior distribution of the regression coefficients at the current iteration of MOBO model; the respective prior distribution is taken as the posterior distribution of the regression coefficients at the previous iteration. The BMLR has been implemented using a *Markov Chain Monte Carlo* (MCMC) approach (no. of Markov chain = 4, no. of warmup iter/chain = 1000, max. iter/chain = 10000) and using the *Gibb's sampling* (GB) algorithm to approximate the posterior distribution of regression coefficients.

*Model 5*: With increasing complexity by adding also square and the pairwise interaction terms of the predictor design variables, we define the quadratic or SOP model as follows:

$$\widehat{\mu_k}(Y_1|R, L, t) = \hat{\beta}_{0,1,k} + \hat{\beta}_{1,1,k}R + \hat{\beta}_{2,1,k}L + \hat{\beta}_{3,1,k}t + \hat{\beta}_{4,1,k}R^2 + \hat{\beta}_{5,1,k}L^2 + \hat{\beta}_{6,1,k}t^2 + \hat{\beta}_{7,1,k}RL + \hat{\beta}_{8,1,k}Lt + \hat{\beta}_{9,1,k}Rt \tag{26}$$

$$\widehat{\mu_k}(Y_2|R, L, t) = \hat{\beta}_{0,2,k} + \hat{\beta}_{1,2,k}R + \hat{\beta}_{2,2,k}L + \hat{\beta}_{3,2,k}t + \hat{\beta}_{4,2,k}R^2 + \hat{\beta}_{5,2,k}L^2 + \hat{\beta}_{6,2,k}t^2 + \hat{\beta}_{7,2,k}RL + \hat{\beta}_{8,2,k}Lt + \hat{\beta}_{9,2,k}Rt \tag{27}$$

*Models 6 & 7*: For the previous models, the errors are assumed independent. Thus, we considered the Support Vector Machine Regression model (SVMR) and the Gaussian process model (GPM) for the case studies where the errors are dependent. These models are altogether different from the earlier stated models, having different statistical approaches in formulation. SVMR and GPM are considered nonparametric techniques because they rely upon kernel functions. SVMR is a popular method in the domain of machine learning and widely used both in classification and regression problems for complex data [58], [59]. SVMR has the flexibility to define how much error is acceptable in the model and fit the data accordingly. Here, the objective is to minimize the sum of $l^2$ norm of the co-efficient vector, $\boldsymbol{\beta}$ and the mean $\varepsilon$-insensitive error, $|\,.\,|_\varepsilon$ of all the data. Thus, the optimization problem can be stated as:

$$\min \frac{1}{n}\sum_{j=1}^{n}\left(\left|y_j - f(\boldsymbol{S}_j)\right|_\varepsilon\right) + \|\boldsymbol{\beta}\|^2 \tag{28}$$

$$\left|y_j - f(\boldsymbol{S}_j)\right|_\varepsilon = |z|_\varepsilon = \begin{cases} 0 & if\ |z| < \varepsilon \\ |z| - \varepsilon & otherwise \end{cases} \tag{29}$$

$$f(\boldsymbol{S}_j) = \boldsymbol{S}_j \boldsymbol{\beta} + b;\ \boldsymbol{S}_j = [R, L, t]_j\ ; j = 1\!:\!n \tag{30}$$

where $\boldsymbol{S}_j$ is the $j^{th}$(row) sampled designs used for fitting the model; $y_j$ is the $j^{th}$(row) true output value; $b$ is the bias, $\varepsilon$ is the error margin and $n$ is the number of training sampled data (rows). The above eqns. (28-30) are the primal formulation where the primal variable is the co-efficient vector $\boldsymbol{\beta}$, for which the dual form is eqns. (31-32), which have been solved using the algorithm Sequential minimal optimization (SMO) [60]. The dual form of (28) can be written as, which is minimized w.r.t the Lagrange multipliers:

$$\min \varepsilon \sum_{j=1}^{n}(\alpha_j^+ + \alpha_j^-) + \sum_{j=1}^{n} y_j(\alpha_j^+ - \alpha_j^-) + \frac{1}{2}\sum_{i=1}^{n}\sum_{j=1}^{n}(\alpha_i^+ - \alpha_i^-)(\alpha_j^+ - \alpha_j^-)G(\boldsymbol{S}_i, \boldsymbol{S}_j) \tag{31}$$

$$\text{subject to } \sum_{j=1}^{n}(\alpha_j^+ - \alpha_j^-) = 0,\ 0 \leq \alpha_j^+, \alpha_j^- \leq C \tag{32}$$

where $\alpha_j^+, \alpha_j^-$ are the $j^{th}$(row) non-negative Lagrange multipliers (dual variables); $G(\boldsymbol{S}_i, \boldsymbol{S}_j)$ is the kernel function between the sampled design, $\boldsymbol{S}$. $C$ is a user defined constant, which balance between model complexity and the approximation error. The primal variables, $\boldsymbol{\beta}(\alpha_j^+, \alpha_j^-)$ can be found as the linear combination of the training sampled designs as eqn. (33).

*chris.hoyle@oregonstate.edu

$$\boldsymbol{\beta}(\alpha_j^+, \alpha_j^-) = \sum_{j=1}^{n}(\alpha_j^+ + \alpha_j^-)\boldsymbol{S}_j \tag{33}$$

Finally, the estimated scores for the objectives $(Y_1, Y_2)$ for a new design $X$ are of the form:

$$\widehat{\mu_k}(Y_1|X) = \sum_{j=1}^{n}\hat{\alpha}_{1,j,k}\boldsymbol{G}_k(\boldsymbol{S}_j, X) + \hat{b}_{1,k} \tag{34}$$

$$\widehat{\mu_k}(Y_2|X) = \sum_{j=1}^{n}\hat{\alpha}_{2,j,k}\boldsymbol{G}_k(\boldsymbol{S}_j, X) + \hat{b}_{2,k} \tag{35}$$

where $\hat{\alpha}_{i,j,k} = \alpha_{i,j,k}{}^+ - \alpha_{i,j,k}{}^-$ is the difference between the two non-negative $j^{th}$(row) Lagrange multipliers for the fitted SVMR model, estimating $i^{th}$ objective of the $k^{th}$ iteration of MOBO model; $\boldsymbol{G}_k(\boldsymbol{S}_j, X)$ is the Kernel function between each sampled design and the new design. In this paper, we considered the kernels as 'Gaussian' and 'linear' for the benchmark problem and the thin tube design, respectively; $\hat{b}$ is the bias estimate. It is to be noted that the

*chris.hoyle@oregonstate.edu

SVMR can be efficient with different kernel functions or tuning of error margin $\varepsilon$. The proposed design architecture has the flexibility to modify SVMR models at any stage of the iteration $k$ of nested MOBO, if required. However, for the sake of simplicity in the comparison, we have avoided that in this case study.

Like the SVMR, GPM is another tool extensively used in machine learning applications [61], [62]. GPM is attractive because of its flexible non-parametric nature and computational simplicity and is, therefore, generally applied within a Bayesian framework, which offers valid estimation and uncertainties in our prediction of function values in nonlinear black-box optimization problems. In our earlier work, we have used Gaussian process, fitted inside a Bayesian optimization framework, in the classification problem as well [55]. The general form of the GP model, given the matrix of designs $\boldsymbol{S}$ for fitting, is as follows:

$$Y(\boldsymbol{S}) = \boldsymbol{S}^T\boldsymbol{\beta} + z(\boldsymbol{S}) \tag{36}$$

where $\boldsymbol{S}^T\boldsymbol{\beta}$ is the 2nd order polynomial regression model. The term $z(\boldsymbol{S})$ is a realization of a correlated Gaussian Process which is defined as follows:

$$z(\boldsymbol{S}) \sim GP\left(E[z(\boldsymbol{S})], cov(\boldsymbol{S}^i, \boldsymbol{S}^j)\right); \boldsymbol{S}_j = [R, L, t]_j; j = 1:n \tag{37}$$

$$E[z(\boldsymbol{S})] = 0, cov(\boldsymbol{S}^i, \boldsymbol{S}^j) = \sigma^2 R(\theta, \boldsymbol{S}^i, \boldsymbol{S}^j); i, j = 1:n \tag{38}$$

$$R(\theta, \boldsymbol{S}^i, \boldsymbol{S}^j) = \exp\left(-\sum_{m=1}^{p} \theta_m \left(s_m^i - s_m^j\right)^2\right); \tag{39}$$

$$\theta_m = \min_\theta |R|^{1/n}\sigma^2 \tag{40}$$

Where $R(\boldsymbol{S}^i, \boldsymbol{S}^j)$ is the spatial correlation function; $|R|$ is the determinant of $R$; $\sigma^2$ is the overall scale parameter and $\theta_m$ is the correlation length parameter in dimension $m$ of $p$ dimension of $x$. The bound of $\theta_m$ is considered as $[1e^{-1}, 20]$ with starting value as 10, as suggested in [63]. These are termed as the hyper-parameters of GP model. The optimal $\theta_m$ is found as solving eqn. 40. In our model, we have used a Gaussian Spatial correlation function which is given as eqn. 39. The estimated score of the objectives $(Y_1, Y_2)$ for a new design $\boldsymbol{X}$ are of the form:

$$\widehat{\mu_k}(Y_1|\boldsymbol{X}) = \boldsymbol{X}^T\widehat{\boldsymbol{\beta}}_{.,1,k} + \boldsymbol{r}(\boldsymbol{X})^T\mathcal{R}^{-1}(Y_1(\boldsymbol{S}) - \boldsymbol{S}^T\widehat{\boldsymbol{\beta}}_{.,1,k}) \tag{41}$$

$$\widehat{\mu_k}(Y_2|\boldsymbol{X}) = \boldsymbol{X}^T\widehat{\boldsymbol{\beta}}_{.,2,k} + \boldsymbol{r}(\boldsymbol{X})^T\mathcal{R}^{-1}(Y_2(\boldsymbol{S}) - \boldsymbol{S}^T\widehat{\boldsymbol{\beta}}_{.,2,k}) \tag{42}$$

$$\boldsymbol{S} = [\boldsymbol{S}_1, \ldots, \boldsymbol{S}_n]^T; \boldsymbol{Y}_q = [y_{q,1}, \ldots, y_{q,n}]^T; \mathcal{R}^{-1} = R(\theta, \boldsymbol{S}^i, \boldsymbol{S}^j)^{-1}; i, j = 1:n;$$

$$\boldsymbol{r}(\boldsymbol{X}) = [R(\theta, \boldsymbol{S}^1, \boldsymbol{X}), \ldots, R(\theta, \boldsymbol{S}^n, \boldsymbol{X})]^T$$

where $\boldsymbol{r}(\boldsymbol{X})$ is the vector of correlation between each sampled design in $\boldsymbol{S}$ and new design $\boldsymbol{X}$; $Y_q(\boldsymbol{S})$ is the vector of true responses ($q^{th}$ objectives) for the respective sampled designs in matrix $\boldsymbol{S}$; $\boldsymbol{S}^T\widehat{\boldsymbol{\beta}}_{.,q,k}$ is the estimated design $q^{th}$ output (objective) function matrix of all the sampled designs $\boldsymbol{S}$; $\mathcal{R}$ is the Gaussian spatial correlation matrix between the sampled designs,

*chris.hoyle@oregonstate.edu

$\mathcal{S}$. It is to be noted the matrix $\mathcal{R}$ is symmetric and positive definite (from the bounds on $\theta_m$) and therefore, matrix $\mathcal{R}^{-1}$ exists and is also symmetric. Thus, as the errors are dependent, the variability in the estimation of any new design which is nearer to the sampled design (used in fitting the model) is lesser than any new design which is farther from the sampled design.

## 5. Results

In this section, we are going to present and discuss the results of comparing the proposed nested weighted Tchebycheff MOBO model with other design architectures at convergence with respect to three major performance criteria. Those are as follows: 1) to maximize the overall prediction accuracy of utopia values, 2) to minimize the number of expensive function evaluations until convergence of MO-BO and 3) to maximize the overall accuracy of converged to true Pareto optimal solutions. We consider weighting factors on objectives distance and cost functions as $w_1 = [0, 0.1, ..., 1]$ and $w_2 = 1 - w_1$ respectively. We used the DACE package [63] in MATLAB for the regression and surrogate GP models. For fitting other regression models, we have used MATLAB functions as fitlm (for fitting MM, MLR, log-MLR and SOP or Quadratic), bayeslm (for fitting BMLR) and fitrsvm (for fitting SVMR). The full nested weighted Tchebycheff MOBO model with *Expected Improvement* type acquisition function has been coded in MATLAB 2018 and run in a machine with configuration of Windows 10, Intel Processor 3.4 GHz and 16 GB RAM.

### 5.1. Case study 1: Benchmark problem

Firstly, we present the results for the multi-objective benchmark problem as referred in eqns. 8-9.

#### 5.1.1. *Discussion on the proportion of models selected by nested MOBO*

Figure 5 shows the proportion of selecting each of the pre-defined six models for model calibration to estimate the model parameters (utopia) towards multi-objective optimization of the benchmark problem over all iterations until convergence of the MO-BO. The bottom right figure is the total number of models selected across all the weight factors of the objectives. To estimate utopia value for both objectives eqns. 8-9, we could see the model selection varies among models with relatively higher complexities than the other pre-defined models, with high percentage for SVMR (Gaussian kernel) and GPM. As we know both the test functions are highly non-linear or multi-modal, the architecture avoided the selection of simpler models; however, we do see a small proportion of simple linear models at the early stage of iteration. This could be because at the early stage the data is limited and so the true nature of the objectives was not identified, and with sequential sampling of data, the MO-BO architecture has been guided to select better regression models as per fig. 5.

*chris.hoyle@oregonstate.edu

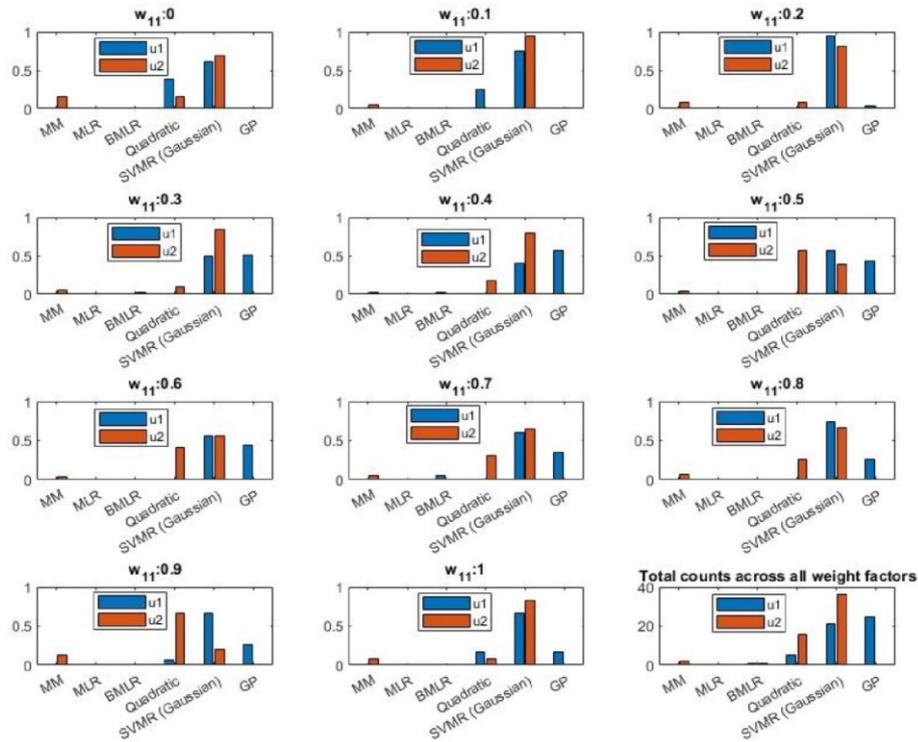

**Figure 5.** Benchmark Problem: Proportion of regression models selected till nested MOBO convergence at various weighting factors for benchmark problem. Here the model portfolio includes: 1. Mean model (MM), 2. Multiple linear regression model (MLR), 3. Bayesian multiple linear regression model (BMLR), 4. Second order polynomial model (SOP) (or quadratic model), 5. Support Vector Machine regression model (SVMR) with Gaussian kernel and 7. Gaussian Process model (GPM).

### 5.1.2. Comparison of different MO-BO architectures

Table 1 shows the overall quantitative measurement of the performance of different MO-BO design architectures across the weighting factors, $w_1$, in terms of the stated performance criteria for the benchmark problem. Figure 6 is the visualization of Table 1 at each weighting factors, $w_1$. To achieve performance criteria 1 and 2, our objective is to minimize the Euclidean norms of utopia prediction and the Pareto-optimal solutions from the true values of utopia and the Pareto-optimal at different weight parameters, and eventually to minimize the mean Euclidean norms across all the weight parameters (Table 1). The true maximum values for both objectives of the benchmark problem are 162.9 and -0.2501 respectively. The true Pareto-optimal are obtained numerically from the exhaustive search using the weighted Tchebycheff method, where the true utopia values are known. With minimizing the mean Euclidean norms, we have focussed on minimizing the standard deviation of the Euclidean norms across different weight parameters. This will ensure that along with the overall minimal solution accuracy, the solution is also less variable across the weights or trade-offs between objectives.

*chris.hoyle@oregonstate.edu

The architectures are summarized in Table 1 below:

- Architecture A: nested MOBO with model selection criteria 1.
- Architecture B: nested MOBO with model selection criteria 2.
- Architecture C: nested MOBO with model selection criteria 1 and 2 (proposed).

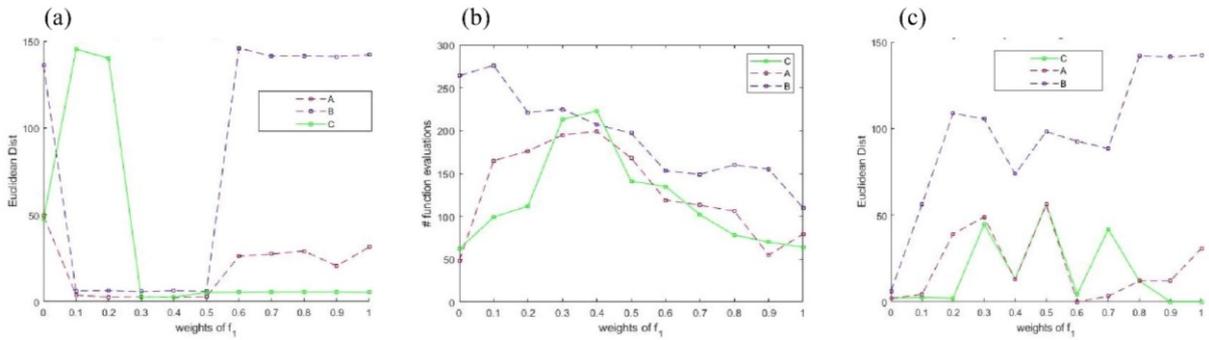

**Figure 6**. Benchmark Problem: (a) Euclidean norms between predicted and true utopia values for $w_1 = [0, 0.1, .., 1]$. (b) Total MO-BO guided func. evaluation till convergence for $w_1 = [0, 0.1, .., 1]$. (c) Euclidean norms between predicted and true pareto-optimal values for $w_1 = [0, 0.1, .., 1]$

| Architecture | | A | B | C |
|---|---|---|---|---|
| Euclidean norm (utopia) | Mean | 18 | 79.87 | 33.73 |
| | St. error | 16.16 | 70.69 | 55.4 |
| Func. eval. | Mean | 129 | 192 | 118 |
| Euclidean norm (Pareto optimal) | Mean | 20.08 | 95.91 | 16.26 |
| | St. error | 20.14 | 40.96 | 20.88 |

**Table 1.** MO-BO design architectures performance comparison for benchmark problem

In analysing the performance of different architectures, we can clearly see (from Table 1) architecture B performs the worst. This intuitively makes sense as the algorithm 2b also depends on the assumption of good overall fitting of the data by the reference model (fitted with full available data). As we did not consider the overall minimal error in fitting over the entire design space, the architecture B is prone to select the regression models only based on the utopia prediction error, where the error can be estimated from bad reference models, thus resulting to an overall lower utopia pareto-optimal solution accuracy. The total expensive function evaluations before convergence are also the highest for architecture B. However, the architectures A and C have an interesting comparison where we see the best (minimum) mean Euclidean norms between the estimated and true utopia across all weight combinations for architecture A, but best (minimum) mean function evaluations and mean Euclidean norms between the MOBO converged and true Pareto-optimal solutions across all weight combinations for the proposed architecture C. With further investigation of C (fig. 6a), we see high mean Euclidean norm of utopia and the respective standard deviation across all weight combinations are due to very high errors for only two weight combinations of $w_1 = 0.1, 0.2$. The accuracy of utopia estimation for other weight combinations are very minimal. However, with the high error in utopia estimation, the final optimal solutions accuracy of C is lower than A for these two weight combinations. Breaking down the utopia estimation accuracy (say for $w_1 = 0.2$), we find the absolute errors of utopia 1 from the true utopia for architectures A and C are 140.1 and 0.08, respectively. The same absolute errors of utopia 2 are 2.37 and 2.42, respectively. Thus for $w_1 = 0.2$, we have much higher preferences on objective eqn.2 than objective eqn. 1. Thus, though the Euclidean norm of utopias for C is higher than A, it has better estimation of utopia 2 for which the objective is highly preferred in the multi-objective optimization. This could be the reason why we see better optimal solutions even with overall higher Euclidean norm of utopias from true values, as the inaccuracy of utopia estimation is comparatively less penalized for the objectives with lower preferences (lower value of weights). We can check this intuitively by putting $w_1 = 0$ in eqn. 9, which will have no effect on the optimal solution accuracy with the increase of the error in utopia 1 estimation, $\hat{\mu}(u_1)$.

*chris.hoyle@oregonstate.edu

### 5.2. Case study 2: Cyclic pressure-temperature loaded thin tube design problem

Here, we present the results for the multi-objective thin tube design problem as referred in eqns. 10-15. Like the earlier problem, we have started with investigating the proportion of each regression model selection to capture the nature of the objectives for estimation of the utopia.

#### 5.2.1. Discussion on the proportion of models selected by nested MOBO

Figure 7 shows the proportion of selecting each of the pre-defined seven models for model calibration to estimate the model parameters (utopia) towards multi-objective optimization of the thin tube design over all iterations until convergence of the MO-BO. To estimate utopia value for objective eqn. 10, we see that model selection varies among log-MLR, BMLR, SOP and GP with higher percentage of log-MLR and SOP. As the weight of objective 1 increases, we see the model selection of utopia 1 shifted from log-MLR to SOP, thus trading off towards the linear model with higher complexities. Furthermore, we do not see any selection of MM, MLR and SVMR (linear kernel). Thus, changing the kernel function affects the selection of the SVMR model significantly as now the model is inefficient to capture any non-linearity of the objectives. This shows the nature of the objective is not fully linear and therefore, relatively simpler linear models considered in this case study are not appropriate here. However, we see a small proportion of Bayesian linear regression model due to its superiority over its frequentist version as it contains the prior information of the regression coefficients. This result agrees with our linearity validation of the objectives when the MO-BO has been calibrated with only MLR models where the assumption was not perfectly met. One interesting observation is when we optimize with the full preference on objective 1 with $w_{11} = 1$ (bottom middle figure), we see almost whole proportion shifted to GP model. This is because of the special case that the multi-objective acquisition function of the MO-BO framework also guides the sampling at the objective 1 utopia region as we are giving importance entirely to minimizing objective 1. Thus, eventually with more sequential sampling in the utopia region, the architecture is flexible to use the benefit of error dependency of Gaussian process as the prediction error of utopia will be much lower. This is the same reason why we see such a high percentage of GPM selection to estimate utopia for objective eqn. 11. We started this MO-BO with the sampling done during the pre-optimization stage [55] where the objective is to locate the unknown creep-fatigue failure constraint (eqn. 12). This region is at the utopia of objective 2 since the minimization of the cost of tube will maximize risk, which eventually is converges towards the creep-fatigue failure constraint. Thus, the architecture has the flexibility to choose regression models for estimation and calibration of the MO-BO based on the starting samples, weighting preferences of the multiple objectives and available sequential sampling guided by acquisition function.

*chris.hoyle@oregonstate.edu

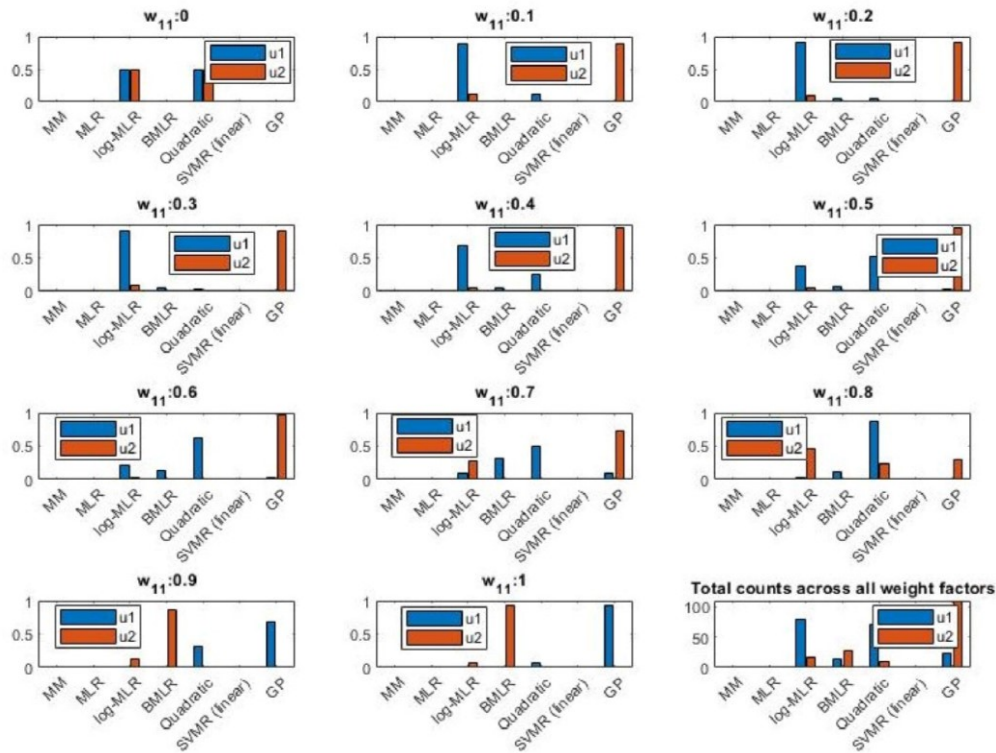

*chris.hoyle@oregonstate.edu

**Figure 7.** Thin tube design problem: Proportion of regression models selected until nested MOBO convergence at various weighting factors for thin tube design problem. Here the model portfolio includes: 1. Mean model (MM), 2. Multiple linear regression model (MLR), 3. Log-transform of multiple linear regression model (log-MLR), 4. Bayesian multiple linear regression model (BMLR), 5. Second order polynomial model (SOP) (or quadratic model), 6. Support Vector Machine regression model (SVMR) with linear kernel and 7. Gaussian Process model (GPM).

*5.2.2. Comparison of different MO-BO architectures*

Table 2 shows the overall quantitative measurement of the performance of different MO-BO design architectures across the weighting factors, $w_1$, in terms of the stated performance criteria for the thin tube problem. Figure 8 is the visualization of the performance of different architectures at each weighting factors, $w_1$. The true maximum values for both objectives of the thin tube problem are 0.1764 and 0.3545, respectively. Similarly, the true Pareto-optimal solutions are obtained from the exhaustive search with true utopia values. The architectures specified in Table 2 is similar to Table 1, but also including two existing architectures D and E where the utopia (unknown parameter) estimation is done with a priory selected regression model. Also, the line connecting black stars * in figs. 10a and 10c are the respective MO-BO model performance without any calibration or iterative estimation of utopia, considering a fixed value of (0, 0): this assumption leads to the worst estimation (as has been addressed in our earlier paper). In this paper, we are drawing comparisons among the architectures where estimation of utopia has been done, but with different procedures. The two existing architectures D and E, where utopia estimation is performed, in Table 2 are defined as below:

- Architecture A: nested MOBO with model selection criteria 1.
- Architecture B: nested MOBO with model selection criteria 2.
- Architecture C: nested MOBO with model selection criteria 1 and 2 (proposed).
- Architecture D: MOBO model integrated with MLR (without ensemble of models).
- Architecture E: MOBO model integrated with BMLR (without ensemble of models).

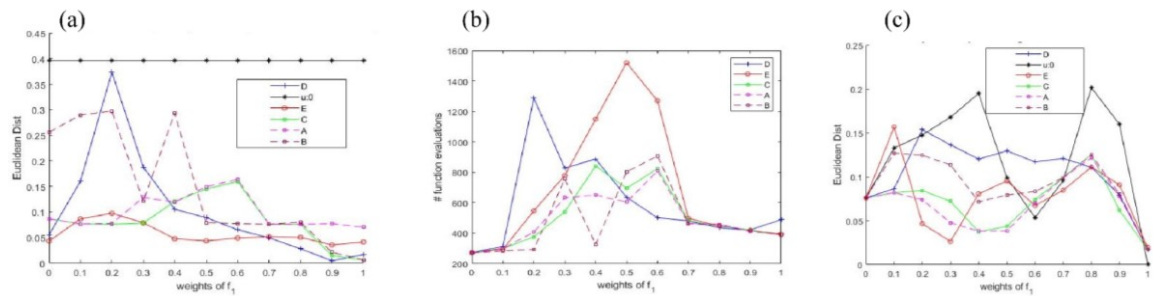

**Figure 8**. Thin tube design problem: (a) Euclidean norms between predicted and true utopia values for $w_1 = [0, 0.1, \ldots, 1]$. (b) Total MO-BO guided func. evaluation till convergence for $w_1 = [0, 0.1, \ldots, 1]$. (c) Euclidean norms between predicted and true pareto-optimal values for $w_1 = [0, 0.1, \ldots, 1]$.

*chris.hoyle@oregonstate.edu

| Architecture | | A | B | C | D | E |
|---|---|---|---|---|---|---|
| Euclidean norm (utopia) | Mean | 0.1 | 0.145 | 0.083 | 0.103 | 0.057 |
| | St. error | 0.034 | 0.115 | 0.047 | 0.106 | 0.021 |
| Func. eval. | Mean | 492 | 487 | 508 | 596 | 690 |
| Euclidean norm (Pareto optimal) | Mean | 0.068 | 0.091 | 0.07 | 0.104 | 0.078 |
| | St. error | 0.03 | 0.033 | 0.03 | 0.038 | 0.039 |

**Table 2.** MOBO design architectures performance comparison for thin tube design

Looking at Table 2, architecture A, C (proposed) and E are the competitive approaches. Comparing our proposed architecture C (considering model selection) with E (considering fixed BMLR), the overall accuracy in utopia estimation, although E has better estimation, are very equivalent as the difference in the accuracy are within the margin or errors. However, E has almost 200 excessive number of function evaluations compared to C which can be viewed as a significant increase of cost when solving an expensive black-box design problem. Next, comparing both A and C (considering model selection with different selection criteria), we get very close results in all the three performance criteria. Thus, we cannot say any architecture clearly supersedes all the others from Table 2. Thus, to measure the performance from the comparison done based on the three criteria, Table 2 has been converted into scores as per Table 3. The score does not only tell the rank of the design architectures, but also provides the measure of how close or far way the performance of an architecture is from the best among them. It is to be noted we calculate the scores only for the mean Euclidean norms, but not for the standard deviation of the same. This is because we give first preference of minimizing the mean norms, as the design architecture having lower mean norms and higher standard deviations is better than higher mean norms and lower standard deviations. In case we have the same mean Euclidean norms for two architectures, the scores for standard deviation of the norms comes into play to break the tie.

From Table 3, we can see existing architecture D is among the lower scores in every criterion, with an overall worst performance (lowest total score) as the linearity assumption was found not to be met. Architecture E has the best scores in predicting utopia (criteria 1); the next best architecture is the proposed architecture C with a score of 70.5. Architecture A is behind architecture E, with half the performance of the best architecture E. In reducing the expensive function evaluation cost (criteria 2), architectures A, B and C clearly outperform architectures D and E, which demonstrates the value of flexibility in selecting regression models with much faster convergence in reaching optimal solutions. Although architecture B has the best scores, the proposed architecture C is close in performance.

28*chris.hoyle@oregonstate.edu

| Architecture | A | B | C | D | E |
|---|---|---|---|---|---|
| **Criteria 1:** Utopia prediction accuracy (Mean Euclidean norm) | 51.1 | 0 | 70.5 | 47.7 | 100 |
| **Criteria 2:** Function evaluation | 97.5 | 100 | 89.7 | 46.3 | 0 |
| **Criteria 3:** Pareto-optimal solution accuracy (Mean Euclidean norm) | 100 | 36.1 | 94.4 | 0 | 72.2 |
| **Total Scores (out of 300)** | **248.6** | 136.1 | **254.6** | 94 | 172.2 |

**Table 3:** Design architecture performance metric (scores). Scores are between 0-100 with 100 being the best and is calculated from individual rows (1,3 and4) of Table 2, following the equation: $s_x = 100 - \left(\frac{x - x_{min}}{x_{max} - x_{min}} * 100\right)$.

With respect to the accuracy of Pareto-optimal solutions (criteria 3), architectures B and D lag behind by significant margins. Both of these architectures also had the worst scores in predicting the utopia, which intuitively make sense as an incorrect prediction of utopia has more likelihood to give incorrect Pareto-optimal solutions. Also, architecture B, as in the benchmark problem, turned out to be worst among A and C. Thus, for both the case studies, we see that although the region of interest is the utopia region, providing some priority on overall good fit in the selection criteria of the regression model is also necessary. Another interesting comparison between A and C is that although the architecture C performance score was much higher in predicting the utopia, it attains a slightly lower score in the accuracy of Pareto-optimal solution than A. As in the benchmark problem, we did a similar investigation and found the reason for this is due to the respective results at $w_1 = 0.3$ (refer fig. 10a, 10c). Thus, in this weight combination, higher penalization should be for the error in estimation utopia 2. However, we see both architectures have the same absolute error of 0.076. Thus, this could be due to another scenario where the utopia prediction is incorrect but along the direction of the weight combination as shown in Appendix fig. A.2. However, an infinite number of such values are possible which is difficult to know, and while incorrect utopia predictions may lead to accurate Pareto solutions (by chance), the focus of this work is to estimate closer to the true utopia to generalize the design architecture for solving any similar problems, not only for this case study. The proposed architecture C is the second best in criteria 3 with only slightly behind the best architecture A. Finally, we see that the proposed architecture C has not been out-performed in any of the performance criteria, but has the best all-around performance with the best score of 254.6. The next best is architecture A. Comparing the standard deviations, we see the same standard deviation for both A and C in the accuracy of Pareto-optimal solutions.

*chris.hoyle@oregonstate.edu

## 6. Conclusion

In this paper, we presented a *nested weighted Tchebycheff multi-objective Bayesian optimization* framework, where the parameter (utopia values) of the acquisition function of the weighted Tchebycheff multi-objective function is estimated from regression analysis in order to calibrate the MOBO for better performance. The utopia estimation is done from the chosen regression model among a pre-defined portfolio of models with various complexities based on a proposed selection criterion. The complete model selection procedure is nested within the MOBO and is formulated to run iteratively as part of the model calibration. The results from the case study of cyclic pressure-temperature load thin tube design problem with two objectives of minimizing risk and cost, shows that the introduction of the flexibility in model selection from portfolio of models for calibration, when we cannot find one best model, has given a much better all-round performance in better estimation of utopia, faster convergence to locate Pareto-optimal solutions and finally better accuracy in locating the Pareto-optimal solutions. The proposed nested MOBO architecture is applicable to any black-box multi-objective optimization problems with minimal or no increase of model complexities with a higher number of multiple objectives; no restriction to define any specific number or families of regression models in the selection procedure; flexibility to compare between two totally different models for utopia estimation (like here with MLR vs SVMR or GPM); and the ability to add, remove or change any regression models other than the defined for this case study either at the start (based on prior expert opinions or information gained from historic analysis) or at any iteration of the MO-BO (with sequential learning on the nature of the black-box objectives). The two selection criteria in choosing the regression model for estimation worked efficiently when coupled together in the nested MOBO design architecture. Although considering only selection criteria 1 in the nested MOBO competes well with the same considering both selection criteria, selection criteria 2 is still important to consider for this problem as the ultimate goal is focused only on efficient prediction of the utopia point.

Based on the analysis from the results discussed, the proposed design architecture can be improved further by investigating the weighting combination between the model selection criteria 1 and 2, which will be addressed in the future. The default setting in this case study is equal preference, which is not optimal. As we understand from comparing architectures A and C in both the case studies, the weighting preference is likely to be higher on selection criteria 1.

*chris.hoyle@oregonstate.edu

Also, another interesting factor which was found during the analysis of the case studies, other than estimation accuracy of utopia (considered in this paper), which effects the Pareto-optimal solution accuracy is the dependency of the penalization of optimal solutions accuracy on the weight preferences on the multiple objectives. Although the focus of this paper is on the weighted Tchebycheff method, the architecture can be easily extended to any other global criterion multi-objective optimization methods where prior knowledge of utopia is required. Finally, the full framework will be implemented in the complex high-dimensional design of a diffusion bonded heat exchanger.

**Acknowledgment**

This research was funded in part by DOE NEUP DE-NE0008533. The opinions, findings, conclusions, and recommendations expressed are those of the authors and do not necessarily reflect the views of the sponsor.

**Competing interests:** There are no conflicts of interest.

*chris.hoyle@oregonstate.edu

*chris.hoyle@oregonstate.edu

*chris.hoyle@oregonstate.edu

*chris.hoyle@oregonstate.edu

**APPENDIX A**

**Figures:**

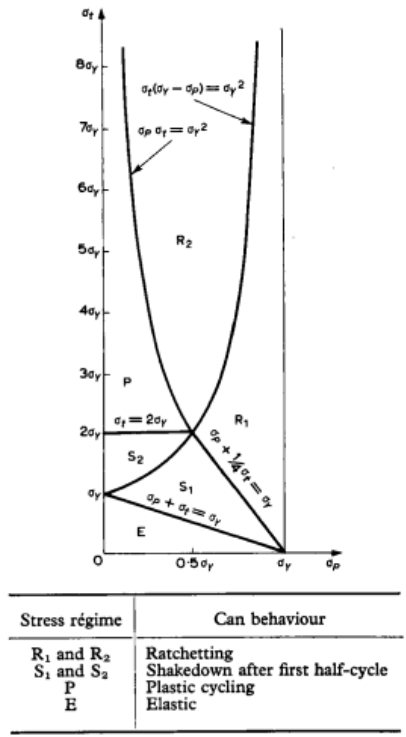

**Figure A.1.** Bree diagram of non-work-hardening material whose yield stress remain unchanged by the change in mean temperature [54]

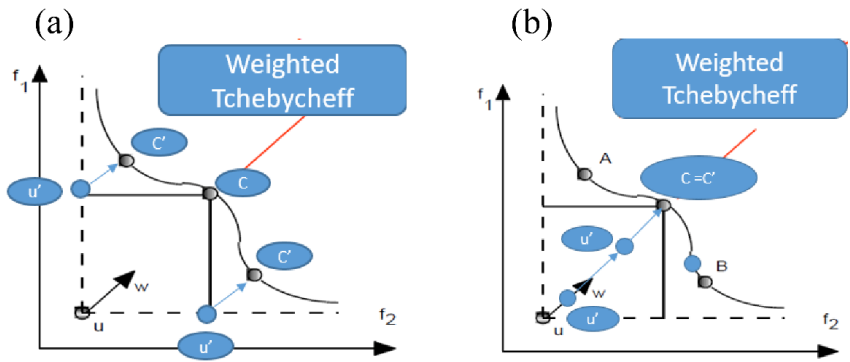

**Figure A.2.** (a) incorrect utopia at $u'$ lead to incorrect pareto-solutions at $C'$ with weight preferences between the objectives, $w$. (b) incorrect utopia at $u'$ along the direction of weight preferences between the objectives, $w$, still provide true pareto-solutions at $C$.

*chris.hoyle@oregonstate.edu